\documentclass[10pt,conference,final]{IEEEtran}
\IEEEoverridecommandlockouts
\usepackage{cite}
\usepackage{amsmath,amssymb,amsfonts}
\usepackage{algorithmic}
\usepackage{graphicx}
\usepackage{textcomp}
\usepackage{xcolor}
\usepackage{subcaption}

\def\BibTeX{{\rm B\kern-.05em{\sc i\kern-.025em b}\kern-.08em
    T\kern-.1667em\lower.7ex\hbox{E}\kern-.125emX}}
\usepackage{multirow}

\usepackage[pagebackref=true,breaklinks=true,colorlinks,bookmarks=false]{hyperref}

\begin{document}

\title{MarineVRS: Marine Video Retrieval System with \\ Explainability via Semantic Understanding}

\author{\IEEEauthorblockN{
    Tan-Sang Ha\IEEEauthorrefmark{1}\textsuperscript{\textsection},
    Hai Nguyen-Truong\IEEEauthorrefmark{1}\textsuperscript{\textsection},
    Tuan-Anh Vu\IEEEauthorrefmark{1} and
    Sai-Kit Yeung\IEEEauthorrefmark{1}\IEEEauthorrefmark{2}%
}
\IEEEauthorblockA{\IEEEauthorrefmark{1}Department of Computer Science and Engineering \\ \IEEEauthorrefmark{2}Division of Integrative Systems and Design (ISD) \\ 
\textit{The Hong Kong University of Science and Technology}, Hong Kong SAR }
}

\maketitle

\begingroup\renewcommand\thefootnote{\textsection}
\footnotetext{Equal contribution}
\endgroup

\begin{abstract}
Building a video retrieval system that is robust and reliable, especially for the marine environment, is a challenging task due to several factors such as dealing with massive amounts of dense and repetitive data, occlusion, blurriness, low lighting conditions, and abstract queries. To address these challenges, we present MarineVRS, a novel and flexible video retrieval system designed explicitly for the marine domain. MarineVRS integrates state-of-the-art methods for visual and linguistic object representation to enable efficient and accurate search and analysis of vast volumes of underwater video data. In addition, unlike the conventional video retrieval system, which only permits users to index a collection of images or videos and search using a free-form natural language sentence, our retrieval system includes an additional Explainability module that outputs the segmentation masks of the objects that the input query referred to. This feature allows users to identify and isolate specific objects in the video footage, leading to more detailed analysis and understanding of their behavior and movements. Finally, with its adaptability, explainability, accuracy, and scalability, MarineVRS is a powerful tool for marine researchers and scientists to efficiently and accurately process vast amounts of data and gain deeper insights into the behavior and movements of marine species.
\end{abstract}


\section{Introduction}

Video retrieval is a challenging task in any domain, but it becomes even more complex in the marine environment due to several factors. These include dealing with massive amounts of dense and repetitive data, which can make it challenging to identify and retrieve specific video clips. Additionally, occlusion caused by marine organisms, blurriness due to water currents, and low lighting conditions all make it difficult to capture clear and high-quality video data. Another significant challenge is dealing with abstract queries, where users may not have specific search parameters in mind, making it difficult to narrow down the search results effectively. Overcoming these challenges requires advanced technologies such as machine learning, artificial intelligence, and computer vision algorithms, as well as a deep understanding of the marine environment and the behavior of marine organisms.

Despite the importance of video retrieval in the marine domain, there is a lack of research in this area. This presents a significant opportunity to develop new methods and technologies to address the unique challenges of retrieving and analyzing underwater video data. Our Marine Video Retrieval System is one such technology that has been developed to revolutionize the way marine researchers, scientists, and experts search and analyze underwater videos.

MarineVRS is a comprehensive video retrieval system that allows users to easily search for and retrieve videos based on specific criteria such as location, time, species, and behavior. It was designed to meet the unique needs of marine researchers and scientists who have to deal with vast amounts of video data collected from various underwater sources. With the help of MarineVRS, they can easily search for and retrieve specific video clips, saving time and resources that would otherwise be spent manually searching through hours of footage.

One of the key strengths of MarineVRS is its advanced search capabilities, which enable users to search for videos based on multiple parameters like location, time, depth, species, and behavior. By integrating state-of-the-art vision-language models like CLIP~\cite{clip} to resolve cross-modality ambiguity, the system achieves high accuracy while maintaining efficiency. This feature makes it possible for researchers and scientists to quickly and efficiently find the videos they need, saving them valuable time and resources that would otherwise be spent in manual searches through hours of footage.

MarineVRS also boasts a range of cutting-edge features that make it a highly effective tool for analyzing underwater video data. For instance, it can output segmentation masks of the target objects, allowing users to easily identify and isolate specific objects in the video footage. This feature enables more detailed analysis and understanding of their behavior and movements, thus helping researchers and scientists to gain insights into the interactions between different marine species and better understand their roles within the ecosystem.

Overall, MarineVRS is a highly adaptive, explainable, and accurate system that is scalable to enormous datasets. It can be customized to meet the specific needs of individual users and organizations, making it highly versatile and flexible. Additionally, MarineVRS is highly explainable, meaning that users can understand how the system arrives at its results, allowing them to make informed decisions about the data. The system has been rigorously tested to ensure that it produces reliable and precise results. In summary, MarineVRS is an essential tool for marine researchers and scientists, providing them with the means to efficiently and accurately process vast amounts of data and gain deeper insights into the behavior and movements of marine species.

\section{Related Work}
In this section, we review some of the recent and fundamental works on image and video retrieval systems, referring image and video segmentation, and marine-related datasets.

\subsection{Retrieval System} Image retrieval systems involve searching for images that match a user's query, while video retrieval systems involve searching for videos that match a user's query. In image retrieval, traditional methods include content-based image retrieval, which searches for images based on their visual features such as color, texture, and shape, and text-based image retrieval, which searches for images based on their associated text metadata. More recent approaches utilize deep learning techniques such as convolutional neural networks (CNNs) to extract features from images and match them with query features using similarity measures. In video retrieval, similar techniques are used, with the additional challenge of dealing with the temporal aspect of videos. Methods include shot boundary detection, keyframe extraction, and feature extraction from video frames or segments. Deep learning techniques such as CNNs and recurrent neural networks (RNNs) can be used to extract spatio-temporal features from videos and match them with query features. Both image and video retrieval systems have numerous applications in various fields such as healthcare, entertainment, and surveillance. Recently, there have been several works that utilize CLIP~\cite{clip} in retrieval systems. For example, CLIP2Video \cite{clip2video} is an end-to-end model for video-text retrieval that leverages a pre-trained image-language model to learn joint representations of images and text. The model consists of two stages: image-text co-learning and temporal relation enhancement and achieves state-of-the-art performance on benchmark datasets. Another work, CLIP4CLIP \cite{clip4clip} is a video retrieval framework that utilizes CLIP~\cite{clip} as a similarity metric to retrieve videos with similar content or style. Recently, there have been several developments in the field of vision-language pretraining for retrieval tasks. For instance, the BLIP \cite{li2022blip} model has shown significant improvements in performance for image and video retrieval tasks, demonstrating the versatility and effectiveness of the CLIP~\cite{clip} model for various retrieval tasks and highlighting its potential for a wide range of applications. 

\subsection{Referring Image and Video Segmentation}
Referring image segmentation is a challenging task that involves generating pixel-wise segmentation masks for referred objects in images based on textual descriptions. Hu et al. (2016)~\cite{huang_referring_2020} first introduced this task, with early works focusing on extracting visual and linguistic features separately from CNNs and RNNs, respectively, and then concatenating them for multi-modal features. Recently, transformer-based multi-modal encoders have been used to fuse visual and linguistic features, capturing early interaction between vision and language information. CRIS leverages CLIP~\cite{clip} and text-to-pixel contrastive learning to improve the compatibility of multi-modal information and cross-modal matching. Referring video object segmentation (RVOS) is a recent extension to video, with URVOS~\cite{urvos} being the first framework followed by PMINet~\cite{pminet} and CITD~\cite{citd} ReferFormer~\cite{referformer} and MTTR~\cite{mttr} are two state-of-the-art works that use transformers to decode or fuse multi-modal features. In this work, we propose a new explainability module that is more efficient and accurate than ReferFormer.

\subsection{Marine-related Datasets}
To better understand marine life and ecosystems, numerous datasets have been created for different tasks such as classification, retrieval, detection, and segmentation. In this section, we will provide a brief review of some recent works that have focused on a marine-related domain that covers a diverse range of marine organisms, such as fish, coral reefs, turtles, whales, and so on.

One such dataset is the Brackish dataset~\cite{pedersen2019brackish}, which contains annotated image sequences of fish, crabs, and starfish captured in brackish water with varying degrees of visibility. This open-access underwater dataset includes 25,613 annotations manually annotated using a bounding box annotation tool. The MOUSS dataset~\cite{mouss2018dataset} is another example, gathered using a horizontally-mounted, grayscale camera positioned 1-2 meters above the sea floor, illuminated solely by natural light. MOUSS seq0 includes 194 images of Carcharhini-formes, while MOUSS seq1 includes 720x480 pixel images of Perciformes, each assigned a species label by a human expert.

The WildFish dataset~\cite{zhuang2018wildfish} is a large-scale benchmark that was developed for the classification task. It consists of 1,000 fish categories and 54,459 unconstrained images. In the field of image enhancement, the Underwater ImageNet dataset~\cite{fabbri2018icra} is composed of subsets of ImageNet~\cite{imagenet} containing photographs taken underwater, with distorted and undistorted sets of underwater images consisting of 6,143 and 1,817 images, respectively.

Another dataset is OceanDark~\cite{porto2019}, which is a novel low-lighting underwater image dataset created for quantitative and qualitative evaluation of proposed image enhancement frameworks. Finally, the Holistic Marine Video (HMV) dataset~\cite{hmv2021} provides a long video simulating real-time marine videos, annotated with scenes, organisms, and actions. It serves as a large-scale video benchmark for multiple semantic aspect annotations and provides baseline experiments for the recognition of marine organism actions, and the detection and recognition of marine scenes.

The latest marine-related dataset is the Marine Video Kit (MVK) dataset~\cite{MVK} comprised of 1379 underwater videos captured at various times and places throughout the year in 36 locations across the globe. The length of the videos ranges from 2 seconds to 4.95 minutes, with a mean and median duration of 29.9 and 25.4 seconds, respectively, for each video. Although the entire duration is slightly longer than 12 hours, the dive time can be up to a thousand hours. We will evaluate our system on this MVK dataset~\cite{MVK}.

\begin{figure*}[!ht]
\includegraphics[width = \textwidth]{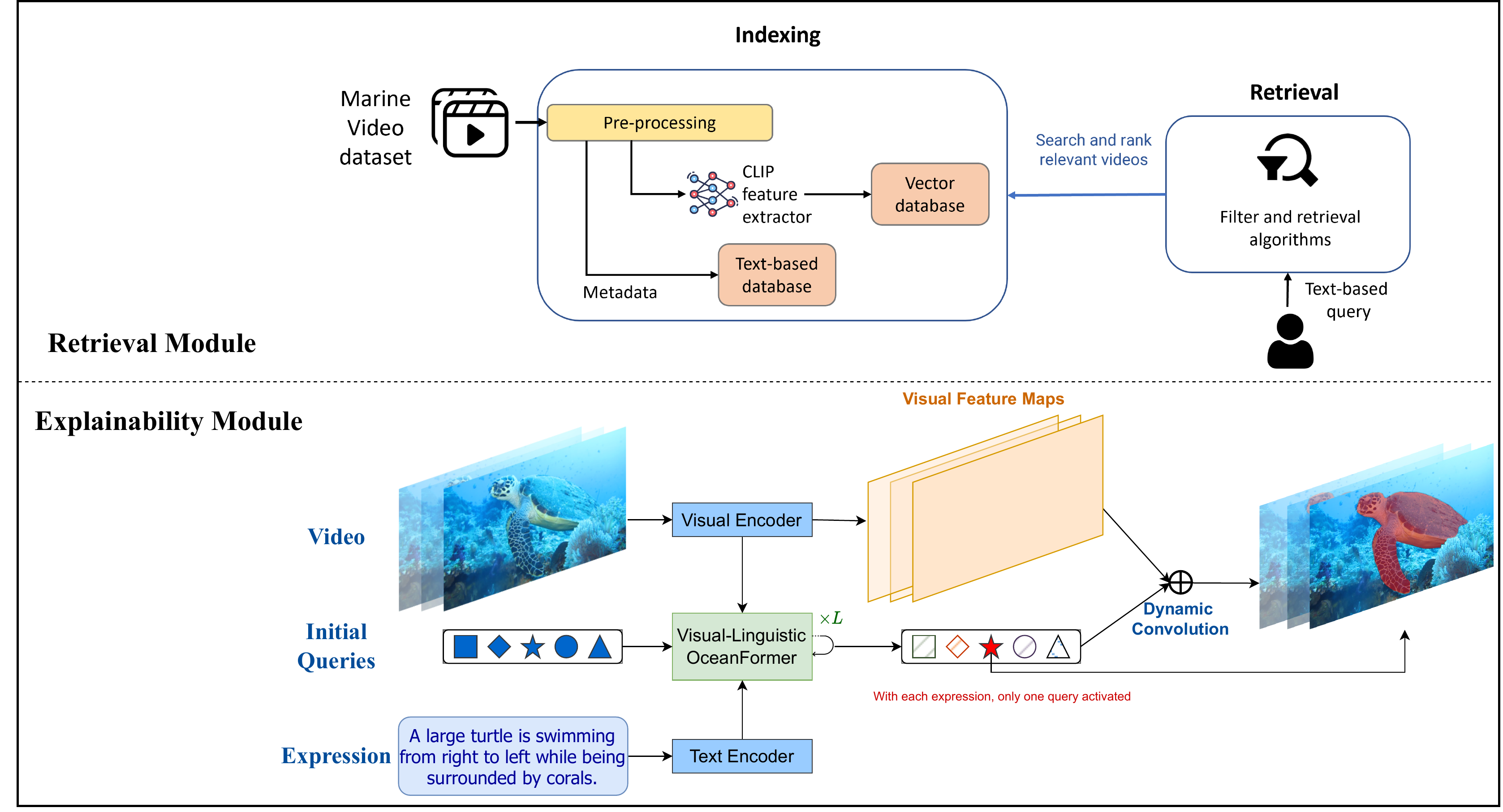}
\caption{\textbf{Overview of Marine Video Retrieval System.} Similar to other retrieval systems, the user's input is a query that describes the requested content in the video database; and then, the system responds with a ranking result of all relevant videos which exist in the database. In addition, by utilizing the explainability module, our system has the capability to embed the segmentation mask of objects or events in the retrieval results. This module facilitates the end-user's job in searching, verifying, and filtering video's content.}
\label{fig1}
\end{figure*}

\section{Method} 
\subsection{Overview}

Figure~\ref{fig1} illustrates the proposed Marine Video Retrieval system which consists of two modules: Retrieval and Explainability. As stated before, our system will mainly focus on the marine domain which is challenging and hasn't been well explored. In the beginning, our Retrieval module will take MVK dataset~\cite{MVK} as input, then we pre-process videos to get CLIP~\cite{clip} features and metadata for indexing. With each text-based query, we will search and rank relevant videos and output them for use as input of the Explainability module. Next, the Explainability module takes a video from Retrieval Module's output as an input for Visual Encoder and a text prompt for CLIP~\cite{clip} Text Encoder. The extracted visual and text features are fused by Multimodal Early Fusion before feeding into Visual-Linguistic OceanFormer to get final segmented masks. Finally, our MarineVRS system will output a set of videos along with their segmented masks or segmented videos which have top similarity scores between the input text query and videos from MVK dataset~\cite{MVK}.

\subsection{Retrieval Module} 
 
The retrieval module of a marine video retrieval system is responsible for returning relevant videos to a user based on their query. In our system, we use the Vision-Language Pre-Training Model CLIP~\cite{clip} to embed both visual and textual data into a shared embedding space, allowing us to calculate similarity scores between the query and videos. However, in order to make this process efficient and scalable, we split the retrieval module into two stages: the indexing stage and the retrieval stage. Figure~\ref{fig1} top illustrates the overview of our retrieval module.

\begin{figure*}[!ht]
\includegraphics[width = \textwidth]{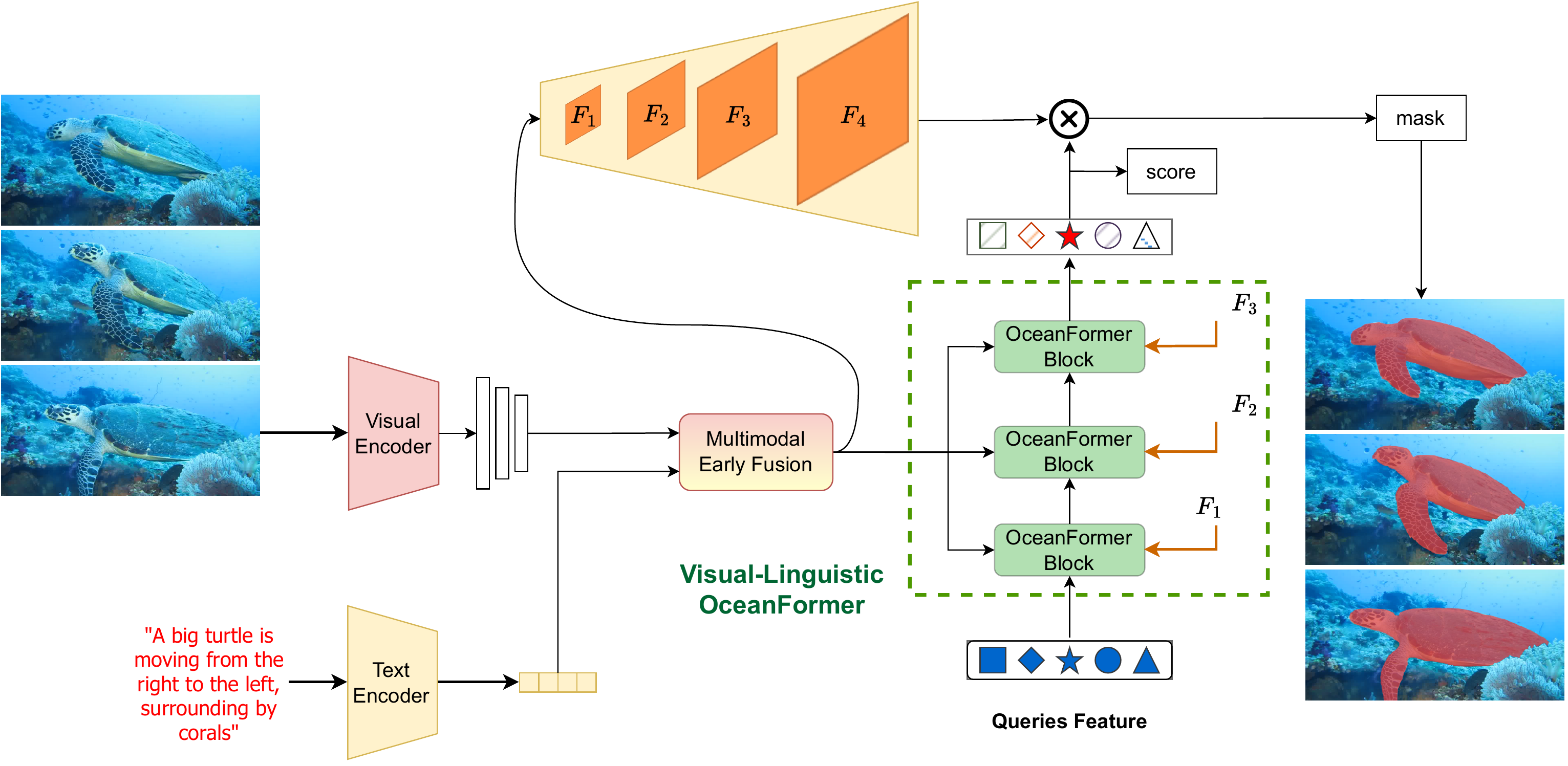}
\caption{Explainability Module. It takes a video from Retrieval Module's output as an input for Visual Encoder and a text prompt for CLIP~\cite{clip} Text Encoder. The extracted visual and text features are fused by Multimodal Early Fusion before feeding into Visual-Linguistic OceanFormer to get final segmented masks.}
\label{fig2}
\end{figure*}

During the indexing stage, we preprocess and store the videos and their metadata in a vector database while maintaining all relevant information. This indexing process allows for faster retrieval and ensures that relevant videos are not lost due to compression or other factors. In the indexing stage of our marine video retrieval platform, we perform several pre-processing steps on the video data to effectively store it in the vector database without losing any information. These pre-processing steps include filtering, normalization, and grouping. First, we filter out blurry images caused by motion or obstruction, because they contain irrelevant information for retrieval purposes. Next, we normalize the images to ensure that they are oriented correctly, as they may have been rotated by multiples of 90 degrees. Finally, we group segments of very similar images caused by stationary viewpoints into one to prevent them from overcrowding the search results. After these pre-processing steps, we use the current state-of-the-art Vision-Language Pre-Training Model CLIP~\cite{clip} to project visual and textual data into the same embedding space.

In the retrieval stage, we use the power of both CPU and GPU to efficiently calculate the similarity scores between the query and the stored videos. This stage involves loading the query and the video database into memory, computing the similarity scores, and then ranking the relevant videos based on their scores. This process can be computationally expensive, but by leveraging the power of both CPU and GPU, we can achieve fast and accurate results.

\subsection{Explainability Module} 
As shown in Figure~\ref{fig2}, the explainability module aims to identify and predict pixel-wise masks of objects within visual data, such as images and videos, based on natural language expressions that refer to these objects. This complex task, known as referring segmentation, presents a higher level of difficulty compared to conventional segmentation tasks, such as semantic and instance segmentation. The main challenges stem from the need to manage open-vocabulary categories and accommodate variations in language syntax, as well as the inherent complexity of the visual data itself.

In the case of the marine dataset, the ambiguous relationships between different objects and their surrounding environment further complicate the task. This ambiguity makes it particularly challenging to accurately distinguish and identify the objects being referred to in the visual data.

Our explainability module consists of three main components: Visual Encoder, Text Encoder, and Visual-Linguistic OceanFormer. The Visual Encoder is a module that extracts visual features from the input images or videos. It can use various techniques such as convolutional neural networks (CNNs) or other computer vision techniques to identify and extract features that are relevant to the task at hand. The Text Encoder is a module that processes the input text data, usually in the form of natural language, to extract meaningful textual features. This module can use various techniques such as recurrent neural networks (RNNs), transformers, or other natural language processing (NLP) techniques to understand text data and extract relevant features. Then, the Multimodal early fusion module is used to combine the visual and textual features extracted by the Visual Encoder and Text Encoder, respectively, at an early stage of the processing pipeline. This can help to create a more integrated representation of the data that incorporates both visual and textual information, potentially improving the model's performance and reducing the risk of information loss. By combining visual and textual features at an early stage, the explainability module can create a more informative query to align vision and language features and indicate the objects referred to by text in the images or videos. This approach could enhance the ability of the module to identify important visual cues and language cues that contribute to the model's decision-making process and provide insights into how the model arrived at its predictions. 

The resulting multimodal features could then be fed into the Visual-Linguistic OceanFormer module for further processing and refinement. It uses a transformer-based architecture to create informative queries that help align the vision and language features and identify the objects referred to by the text in the images or videos. This module can help the model to better understand the relationships between visual and textual information and produce more accurate predictions. These query features, together with the last visual feature map, produce the high-resolution segmentation mask. During inference, the query with the highest confidence score is selected as the target object for the final output. Overall, the explainability module is designed to provide insights into how the model arrived at its predictions. By extracting visual and textual features and combining them in a meaningful way, the module can help identify important visual cues and language cues that contribute to the model's decision-making process. This can be valuable in helping users to understand and interpret the model's output, as well as identify potential biases or errors in the model.

The model is first trained and evaluated on four well-established public datasets: RefCOCO~\cite{refcoco}, RefCOCO+~\cite{refcoco}, G-Ref~\cite{gref}, and Referring YoutubeVOS~\cite{urvos}. These datasets provide a solid foundation for the model to learn and understand a wide range of object categories and language structures. After this initial training phase, the model is fine-tuned on a smaller selection of marine videos, which helps it adapt to the unique challenges presented by the marine environment. Once the fine-tuning process is complete, the model is integrated into our platform, where it can effectively segment objects in visual data based on natural language expressions. This approach enables the explainability module to deliver accurate and reliable object identification and segmentation, even in the challenging and complex context of marine environments.

\textbf{Loss function}: Our goal is to generate a small group of $N$ predictions, from which we will choose the best one as the final object. We use the instance-matching strategy to supervise candidate instances during network training. The predicted set is denoted by $\hat{y} = \left\{\hat{y_i}\right\}_{i = 1}^{N}$, which consists of two components: $\hat{p}_i \in \mathbb{R}$ indicating the probability of the instance corresponding to the referred object, and $\hat{s}i \in \mathbb{R}^{T \times H \times W}$ as the segmentation mask. We assume that the image is equivalent to a video with a duration of $T=1$, and the ground-truth object is represented as $\hat{s}_i \in \mathbb{R}^{T \times H \times W}$ since there is only one referred object. To train the network, we minimize the matching cost $\mathcal{L}_{\text{match}}$ by selecting the best prediction $i$-th from $N$ candidates.The matching cost measures the quality of the match between two sets of objects, where each object has a query feature and a mask prediction. It's calculated using three loss functions: $\mathcal{L}_{cls}$, $\mathcal{L}_{mask}$, and $\mathcal{L}_{dice}$. The first loss function, $\mathcal{L}_{cls}$, compares the predicted and actual probabilities that a query feature corresponds to the referred object using Binary Cross-Entropy loss. The second loss function, $\mathcal{L}_{mask}$ supervises the mask prediction and measures the difference between the predicted and actual masks at the per-pixel level using Cross-Entropy loss. The third loss function, $\mathcal{L}_{dice}$ improves the dice score by measuring the overlap between the predicted mask and the ground truth mask for the referred object. The goal is to minimize the matching cost by finding the best prediction from $N$ candidates while ensuring that other queries don't represent the referred object. We use the same method as Mask2Former~\cite{mask2former} to calculate this loss, and the corresponding loss function coefficients are denoted as $\gamma_{cls}$, $\gamma_{mask}$, and $\gamma_{dice}$. The objective is to minimize the $\mathcal{L}_{\text{match}} = \gamma_{mask}\mathcal{L}_{mask} + \gamma_{dice}\mathcal{L}_{dice}$ of the matched query and prevent other queries from representing the referred object. Therefore, the matching cost and loss function is defined as:

\begin{equation}
    \mathcal{L}(y, \hat{y}, i) = \mathcal{L}_{match}(y, \hat{y}_i) + \sum_{\substack{j = 1 \\ j \not=i}}^{N}{\gamma_{cls}\mathcal{L}_{cls}(\hat{p}_j, 0).}
\end{equation}

During the inference, the explainability module generates a set of predictions, with each prediction consisting of a query feature and a segmentation mask. The query with the highest confidence score is selected as the target object for the final output. However, due to the memory limitations of the current system (such as GPU memory), it may not be possible to process an entire video at once. Therefore, to accommodate this constraint, the explainability module processes each video in chunks of 32 frames at a time. For each chunk, the module generates a set of predictions and outputs the segmentation mask of the referred object (if any). This approach allows the explainability module to provide insights into the model's decision-making process and identify the objects referred to by the text in the video, while also managing the computational and memory requirements of the system.

\section{Experiments}

In this section, the implementation details and experimental results of the proposed system are provided.

\subsection{Implementation Details}

Our whole system is deployed and tested on a local PC with the following specifications, CPU Intel Xeon Silver 4316 (30M Cache 2.30 GHz), GPU NVIDIA RTX 3090 24GB, 64GB for RAM, and 2TB for storage.

For the Explainability module, our implementation is based on the PyTorch framework and involves freezing both the CLIP Visual and Text Encoder during the training process. To initialize the visual encoder and pixel decoder, we leverage the pre-trained Mask2Former model~\cite{mask2former}. During training, we resize the images to have a short side of 480. We set the coefficients for the losses as $\gamma_{cls} = 2$, $\gamma_{mask} = 5$, and $\gamma_{dice} = 5$, with the feature dimension $C$ set to $256$. We use the AdamW optimizer~\cite{adamw} to train the network for $100,000$ iterations on the RefCOCO, RefCOCO+, and G-Ref datasets. The initial learning rate is set to $0.0001$ and is reduced by a factor of $0.1$ at the $70,000$th iteration.

For the Retrieval module, we employed the novel vector database Milvus\cite{2021milvus,2022manu} to efficiently index and search large-scale video datasets, providing faster and more accurate retrieval of relevant content. To demonstrate our system, we used the user-friendly platform Gradio~\cite{abid2019gradio}, which simplified the user interface and allowed us to showcase our system's performance to wider users. As our system expands, we can scale up our infrastructure to accommodate more users without sacrificing performance.

\subsection{Quantitative Results}

\begin{table}[!htb]
\centering
\caption{Comparison with the state-of-the-art methods on Ref-COCO, Ref-COCO+, G-Ref and Ref-YoutubeVOS datasets on IoU and $\mathcal{J} \& \mathcal{F}$. ``-'' denotes that the result is not available.}
\resizebox{\columnwidth}{!}{%
\begin{tabular}{|c|c|c|c|c|}
\hline
                         & \textbf{RefCOCO} & \textbf{RefCOCO+} & \textbf{G-Ref} & \textbf{Ref-YoutubeVOS} \\ \cline{2-5} 
\multirow{-2}{*}{\textbf{Method}} & IoU     & IoU      & IoU   & $\mathcal{J} \& \mathcal{F}$            \\ \hline\hline
CRIS         \cite{cris}            & 70.47   & 62.27    & 59.87 & -              \\
LAVT                    \cite{lavt} & 72.73   & 62.14    & 61.24 & -              \\ 
ReferFormer  \cite{referformer}        &     71.1 &64.1 &64.1 & 55.66 \\ 
\textbf{Ours}                      & \textbf{76.15}   & \textbf{65.02}    & \textbf{66.87} & \textbf{59.15}          \\ \hline
\end{tabular}%
}

\label{tab:my-table}
\end{table}

Our method consistently surpasses other state-of-the-art methods across all datasets. In the RefCOCO dataset, our approach exhibits improvements of 8.07\%, 4.71\%, and 7.11\% over CRIS, LAVT, and ReferFormer, respectively. For the RefCOCO+ dataset, our method achieves performance boosts of 4.41\% compared to CRIS, 4.66\% over LAVT, and a modest 1.43\% over ReferFormer. Turning to the G-Ref dataset, our technique excels with an 11.72\% advantage over CRIS, 9.2\% over LAVT, and a 4.31\% lead against ReferFormer. Lastly, in the Ref-YoutubeVOS dataset, our method exhibits a substantial 6\% improvement over ReferFormer. These results emphasize the effectiveness and robustness of our method in comparison to other approaches in diverse benchmarks.

\subsection{Qualitative Results}


%
\begin{figure*}[!htb]
    \centering
    \subfloat[Example 1: Search for shark]{{\includegraphics[width = .49\textwidth]{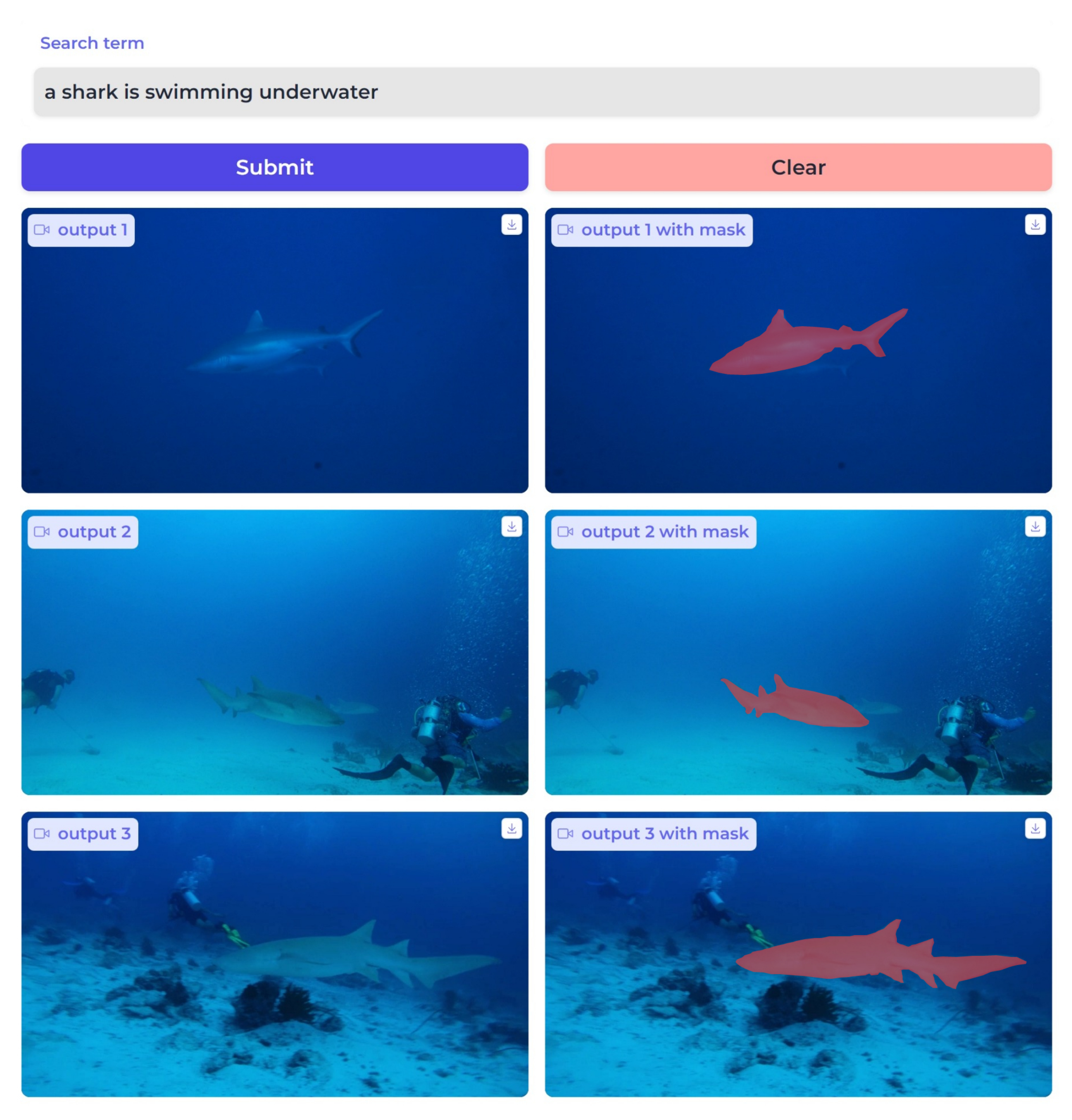}}}%
    \subfloat[Example 2: Search for turtle]{{\includegraphics[width = .49\textwidth]{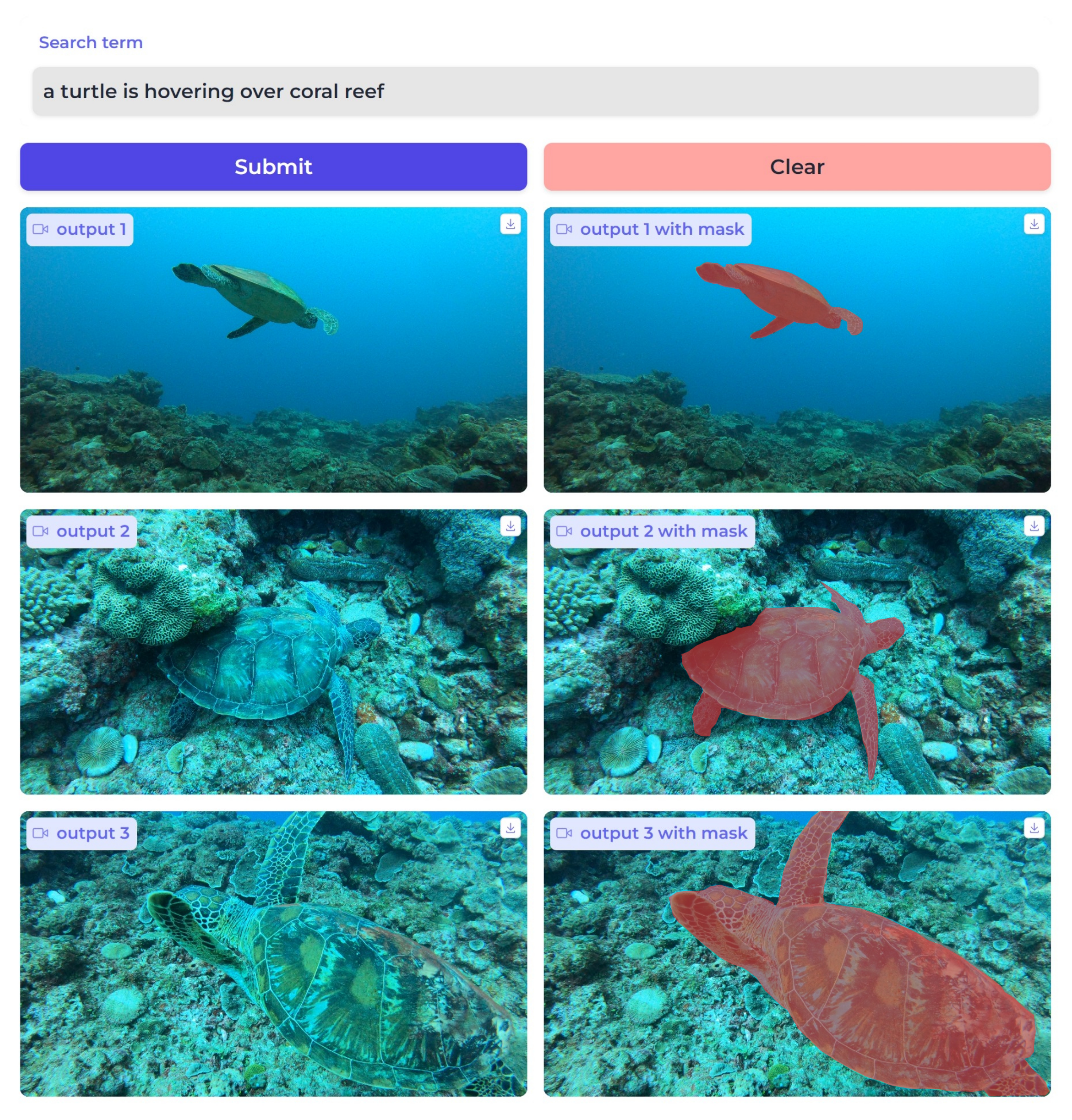}}}%
    \caption{Our retrieval system's performance and usability are demonstrated by a prototype application to prospective users and stakeholders.}%
    \label{fig-search-with-mask}
\end{figure*}

Figure~\ref{fig-search-with-mask} shows some examples from our MarineVRS system. With text prompts as input, our system will search and retrieve top results (in these examples, we selected the top 3 results) that relate to an object recognition and tracking system that is capable of detecting and tracking marine species' movement in an underwater environment. The use of object masks in the output videos provides a powerful tool for achieving explainability and visual evidence of the system's performance. This level of explainability can be particularly beneficial in situations where the accuracy and reliability of the system's output are critical, such as in the field of marine biology where researchers may be using the system to monitor the behavior of different marine species. Incorporating object masks in the output videos is an excellent method to provide a visual representation of the system's performance and to help users comprehend how it detects and tracks objects in real-world settings. The use of object masks can help viewers see precisely where the system identifies marine species and how it tracks their movement over time, allowing for a deeper understanding of the system's functioning. The approach can be particularly useful in domains where accuracy and reliability are critical, such as marine biology.

\subsection{Running time} 
We studied the overhead of our system by providing a breakdown of the end-to-end running time. Overall, our system takes around 1 second to display the retrieved and segmented videos since the system received the text query input. While having impressive results, the Retrieval module of our system relies on a vector data management system - Milvus~\cite{2021milvus} to efficiently index and search large-scale videos on a local PC and thus have more overhead compared to other commercial systems which utilizes a much more complex and expensive data management system and are deployed on powerful servers or clouds. On the other hand, our Explainability module not only achieves state-of-the-art performance but also maintains real-time inference speed ($\sim$ 31 FPS) which is remarkable for any online platform.

\section{Conclusion}
In conclusion, the MarineVRS is a powerful video retrieval system designed to tackle the challenges of analyzing underwater video data. By integrating state-of-the-art methods for visual and linguistic object representation, the MarineVRS provides efficient and accurate search capabilities for vast volumes of data. Furthermore, MarineVRS is highly adaptable, explainable, accurate, and scalable, making it an essential tool for marine researchers and scientists to gain deeper insights into the marine environment.

Our system has a wide range of applications in various computer vision tasks, such as searching and browsing huge video archives, indexing and archiving multimedia, surveillance, and monitoring for underwater species or coral conservation, and so on. In addition, as deep learning requires a substantial amount of labeled training data, it is intriguing how an image or video retrieval might be utilized to help prepare training datasets for improving deep model training.

In the future, we intend to enlarge our investigation of potential solutions to improve the selection and re-ranking of videos in our system. Specifically, we plan to utilize the ranking of videos received from a selection network to improve the identification of relevant matches for video queries. By reducing the need to scan through vast volumes of data, this strategy will increase the effectiveness and efficiency of our retrieval system. Further improving the system's scalability without sacrificing retrieval performance, we intend to look into new network architectures for both modules. Our goal is to develop a very flexible and effective video retrieval system that can manage enormous volumes of data while giving users accurate and quick results. By following these research directions, we intend to create a system that is more complete and capable of meeting the changing requirements of marine researchers and scientists.

\section*{Acknowledgment}
\addcontentsline{toc}{section}{Acknowledgment}
This research project is partially supported by an internal grant from HKUST (R9429), the Innovation and Technology Support Programme of the Innovation and Technology Fund (Ref: ITS/200/20FP), and the Marine Conservation Enhancement Fund (MCEF20107).

\bibliographystyle{IEEEtran}
\bibliography{refs}

\end{document}